%% file: main.tex
\documentclass[10pt,twocolumn,letterpaper]{article}

\usepackage{multirow}
\usepackage{cvpr}      
\usepackage{xcolor}

\definecolor{cvprblue}{rgb}{0.21,0.49,0.74}
\usepackage[pagebackref,breaklinks,colorlinks,citecolor=cvprblue]{hyperref}
\def\paperID{\textbf{52}} 
\def\CodalabID{\textit{\textbf{AsheLee}}} 
\def\confName{CVPR}
\def\confYear{2024}

\title{Generalized Few-Shot Meets Remote Sensing: Discovering Novel Classes in Land Cover Mapping via Hybrid Semantic Segmentation Framework}

\author{Zhuohong Li$^{1*}$, Fangxiao Lu$^{1*}$, Jiaqi Zou$^1$, Lei Hu$^1$, Hongyan Zhang$^{1, 2\dag}$\\
$^1$Wuhan University \hspace{+0.7em} $^2$China University of Geosciences\\
\tt\small \{ashelee, fangxiaolu, immortal, hulei.eva\}@whu.edu.cn, zhanghongyan@cug.edu.cn
}
\begin{document}
\maketitle
\input{sec/0_abstract}
\input{sec/1_intro}

\input{sec/2_related_work}
\input{sec/3_Method}
\input{sec/4_Experiment}

\input{sec/5_Conclusion}
{
    \small
    \bibliographystyle{unsrt} 
    \bibliography{main}
}

\end{document}

%% file: sec/0_abstract.tex
\begin{abstract}
Land-cover mapping is one of the vital applications in Earth observation, aiming at classifying each pixel's land-cover type of remote-sensing images.
As natural and human activities change the landscape, the land-cover map needs to be rapidly updated. However, discovering newly appeared land-cover types in existing classification systems is still a non-trivial task hindered by various scales of complex land objects and insufficient labeled data over a wide-span geographic area. In this paper, we propose a generalized few-shot segmentation-based framework, named SegLand, to update novel classes in high-resolution land-cover mapping. 
Specifically, the proposed framework is designed in three parts:
(a) Data pre-processing: the base training set and the few-shot support sets of novel classes are analyzed and augmented;
(b) Hybrid segmentation structure:  Multiple base learners and a modified Projection onto Orthogonal Prototypes (POP) network are combined to enhance the base-class recognition and to dig novel classes from insufficient labels data; (c) Ultimate fusion: the semantic segmentation results of the base learners and POP network are reasonably fused. The proposed framework has won first place in the leaderboard of the OpenEarthMap Land Cover Mapping Few-Shot Challenge. Experiments demonstrate the superiority of the framework for automatically updating novel land-cover classes with limited labeled data.\renewcommand{\thefootnote}{}
\footnotetext{$^*$Indicates equal contribution. 
$^\dag$Corresponding author. The code is open-access at \url{https://github.com/LiZhuoHong/SegLand}} 
\end{abstract}

%% file: sec/1_intro.tex
\section{Introduction}
\label{sec:intro}
\begin{figure}[t]
{
    \begin{minipage}[b]{\hsize}
     \centering
    \includegraphics[width=\linewidth]{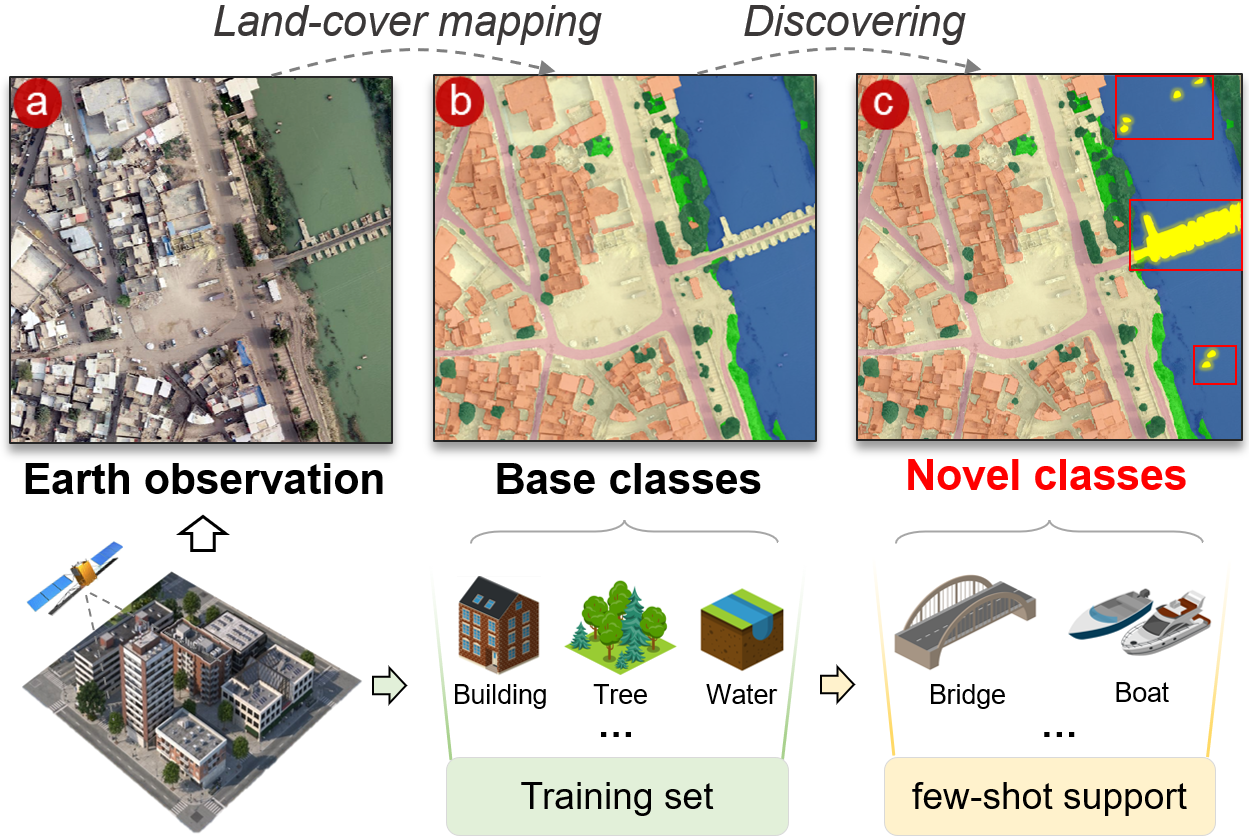}
    \end{minipage}
    } 
    \vspace{-2em}
\caption{ \footnotesize \rmfamily Illustration of discovering novel classes in the general land-cover mapping process. (a) \textbf{Earth observation} provides HR remote-sensing images. (b) \textbf{Base classes} contain sufficient training set. (c) \textbf{Novel classes} suffer from insufficient labeled data (\textit{i.e.,} few-shot support set).}
\label{intro}
\vspace{-1em}
\end{figure}

Land-cover mapping is a particular semantic segmentation problem shared in the fields of Earth observation and computer vision. 
The land-cover map is up-to-date data and should be continuously updated along with landscape changes \cite {robinson2019large}. 
In the past few decades, low/medium-resolution land-cover mapping has made significant progress. Currently, available high-resolution (HR) remote-sensing images ($\le$ 1 meter/pixel) provide an opportunity for finer land-cover mapping \cite{TONG2020111322}. 

With higher spatial resolution, richer land objects that were previously unseen can now be observed. However, discovering novel classes in HR land-cover mapping is still a non-trivial task hindered by the various scales of land objects and also the scarcity of training labels over a wide-span geographic area. The time-consulting and laborious annotation inevitably limited the volume of novel-class labels, which is still a main challenge for updating finer land-cover maps across large-scale areas
\cite{li2022outcome,girard2021polygonal,zhang2022seamless}. 

As shown in Fig. \ref{intro}(a) and (b), general land-cover mapping uses sufficient training data to identify each pixel's land-cover class in remote-sensing images. Such a regular pipeline has dominated many advanced methods for land-cover mapping. No matter the pixel-to-pixel machine learning methods (\textit{e.g.,} random forest \cite{chan2008evaluation}) or deep learning-based methods (\textit{e.g.,} CNN-based \cite{robinson2019large,xu2022luojia} and Transformer-based \cite{wang2022unetformer, wang2022novel}) require enough labeled samples for each land-cover classes during the training process. As a result, numerous large-scale land-cover products have also been produced via similar workflows, such as the sub-meter-level land cover map of Japan \cite{yokoya2024submeter}, the first 1-meter land-cover map of China \cite{li2023sinolc}, and other 10-meter global land-cover maps \cite{van2021esa}. 

With the development of Generalized Few-shot Semantic Segmentation (GFSS), researchers are allowed to discover the novel classes with very few samples and also learn the base land-cover classes with sufficient labeled samples \cite{tian2022generalized}. Although GFSS has shown great success in quickly adapting to novel classes of natural images with few labeled data (\textit{e.g.,} 1-shot and 5-shots), discovering novel classes in remote-sensing images is even more unique and different. 
\begin{enumerate}
    \item \emph{The land objects vary in scale and size. For example, the classes of cropland, forest, and buildings generally have larger sizes. The novel classes such as boats, cars, and bridges have smaller sizes \cite{wang2022cross}.}

    \item \emph{ The objects of base and novel classes are interconnected and easily misclassified. For example, the lakes and ponds (base class) have almost identical attributes to the rivers (novel class). Similar situations exist in different classes of roads, agricultural land, and buildings \cite{xia2023openearthmap}.}  
\end{enumerate}

To address these challenges, we propose a GFSS-based land-cover mapping framework, named SegLand, to discover the novel land-cover classes that appear on the base land-cover maps. Specifically, the framework consists of three main parts: (a) Data pre-processing: the base training set and the few-shot support sets of novel classes are analyzed and augmented;
(b) Hybrid segmentation structure:  Multiple base learners and a modified Projection onto Orthogonal Prototypes (POP) network are combined to enhance the base-class recognition and to dig novel classes from insufficient labels data; (c) Ultimate fusion: the semantic segmentation results of the base learners and POP network are reasonably fused.

%% file: sec/2_related_work.tex
\section{Related Work}
\label{sec:formatting}
\subsection{Generalized Few-Shot Segmentation}
Few-shot semantic segmentation (FSS) performs pixel-wise recognition on novel classes given a few annotated support examples. Despite showing great potential, the FSS approaches clearly rely on overly powerful prior knowledge of unseen/novel classes. GFSS for more practical scenarios is further proposed by Tian \emph{et al} \cite{tian2022generalized} to solve this issue. In particular, GFSS frees the harsh constraint that support and query images must contain the same categories. Furthermore, GFSS aims to recognize novel categories using a few examples without discarding the segmentation accuracy of the base categories.


With these advantages, the GFSS has garnered plenty of attention. CAPL \cite{tian2022generalized} proposed the first attempt to tackle GFSS tasks by using adaptive features to dynamically enrich contextual information, achieving significant performance improvements on both base and novel classes. Nevertheless, the displayed results are skewed toward the base classes and require the assistance of the labeled base classes. Furthermore, BAM \cite{lang2022learning} also attempted to evaluate segmentation performance under a generalized setting. However, the regrettable truth is that the meta-learner cannot be directly used for multi-class GFSS tasks. Fine-tuning \cite{myers2021generalized} is another simple yet effective solution to adapt the models for the GFSS tasks. Unfortunately, the model performs poorly on the base classes or new classes. The latest DIaM (baseline model provided by OEM Few-Shot Challenge) \cite{hajimiri2023strong}, which was rooted in the InfoMax framework with problem-specific biases. Since the training data mainly comes from the base classes, the GFSS mode is inevitably biased towards the base classes. To resolve this difficulty, PCN \cite{lu2023prediction} focused on a fusion strategy for the scores produced by the base and novel classifiers respectively.



The training strategy of the GFSS methods mentioned above is relatively typical, and we can summarize it in two steps: base class learning and novel class updating. However, independent updates can put well-learned features at risk and lead to performance degradation on base classes. POP \cite{liu2023learning} innovatively proposed a new idea of using projection onto orthogonal prototype, which updates features to identify novel classes without sacrificing excessive accuracy on base classes. In this paper, we implement GFSS of remote sensing images using POP as the baseline, and the detailed technical scheme is presented in Section 3.2. 

\subsection{Land-cover labeled data}
Recent approaches have shortened the HR land cover mapping cycle through various efficient methods. Many large-scale HR land-cover maps, such as the submeter-level land-cover map of Japan \cite{yokoya2024submeter} and the first 1-meter land-cover map of China \cite{li2023sinolc} can be produced with Less time and cost spent. However, for a general land-cover mapping pipeline (especially for discovering novel classes), 
Creating sufficient labeled samples is still extremely time-consuming
and expensive \cite{pengra2015global,dong2021high}. In general, current existing land-cover labeled data can be summarized as follows: \\
\textbf{1)  Global-scale low-resolution products:}
From the 1980s to the 2000s, global-scale imagery with low resolution can be captured by SPOT 4, MODIS, and ENVISAT missions. Subsequently, many global land-cover (GLC) products, \textit{e.g.,} 300-m GlobCover, have emerged \cite{defourny2006globcover}.
\\
\textbf{2) Global-scale moderate/high-resolution products:}
From the 2010s to 2020s, owing to the available Landsat and Sentinel imagery with moderate ($\sim$30m) and high ($\sim$10m) resolution, the related research has blossomed. \textit{E.g.,} GlobeLand30 \cite{chen2015global}, FROM\_GLC10 \cite{chen2019stable}, ESA\_WorldCover \cite{van2021esa}.
\\
\textbf{3) Region-scale very-high-resolution datasets:}
In the 2020s, creating VHR datasets for deep learning research has become a hotspot and current VHR land-cover datasets
(\textit{e.g.,} 2.1-m Hi-ULCM and 2.4-m PKU-USED) for China are
generally regional-scale (typically covering a few cities) \cite{huang2020high}.
\subsection{Land-cover mapping with limited samples}
Insufficient labeled data is a common issue shared in the field of Earth observation and computer vision. Especially, land-cover mapping task is usually conducted in a very large-scale geographic area. How to solve the problem of scarcity of labeled data has become an important issue in this field \cite{li2022breaking}. Currently, there are many advanced approaches to save laborious annotation by conducting the mapping process with limited labeled samples.\\
\textbf{1)  Mining inexact labeled data:}
Due to the scarcity of accurate training labels, prior studies focused on mining inexact labeled data (\textit{e.g.,} rough and noisy labels) to supervise the land-cover mapping process. For example, the winner approach of the 2021 IEEE Data Fusion Contest (DFC) \cite{li2022outcome}, the low-to-high network (L2HNet) \cite{li2022breaking}, and Paraformer \cite{li2024learning} were designed to conduct HR land-cover mapping only using LR-labeled data.
\\
\textbf{2)  Mapping with insufficient labels:}
As another vital aspect, numerous approaches focus on using insufficient local labels to conduct land-cover mapping with wider coverage. These methods generally rely on a semi-supervised strategy. For example,  in the 2022 DFC, semi-supervised land-cover mapping was regarded as a vital issue of Earth observation \cite{b9pt-8x03-20, li2022multi}. In the general pipeline, a very large-scale land-cover mapping task should be conducted under the supervision of accurate labels in a few places.   
\\
\textbf{3)  Few-shot learning in remote sensing:}
Recently, FSL has been gradually extended to many fields of remote sensing with satisfactory results. For the semantic segmentation task, the current trend is to use metric learning to distinguish target categories that do not exist in the source domain. For example, the spatial-spectral relation network (SS-RN) \cite{rao2019spatial} and DMCM \cite{hu2023cross}. However, more generalized segmentation scenarios need to be explored, i.e. not just focusing on new classes nor abandoning base class recognition.

%% file: sec/3_Method.tex
\section{Method}
\begin{figure*}[t]
    \centering
    \includegraphics[width=0.9\linewidth]{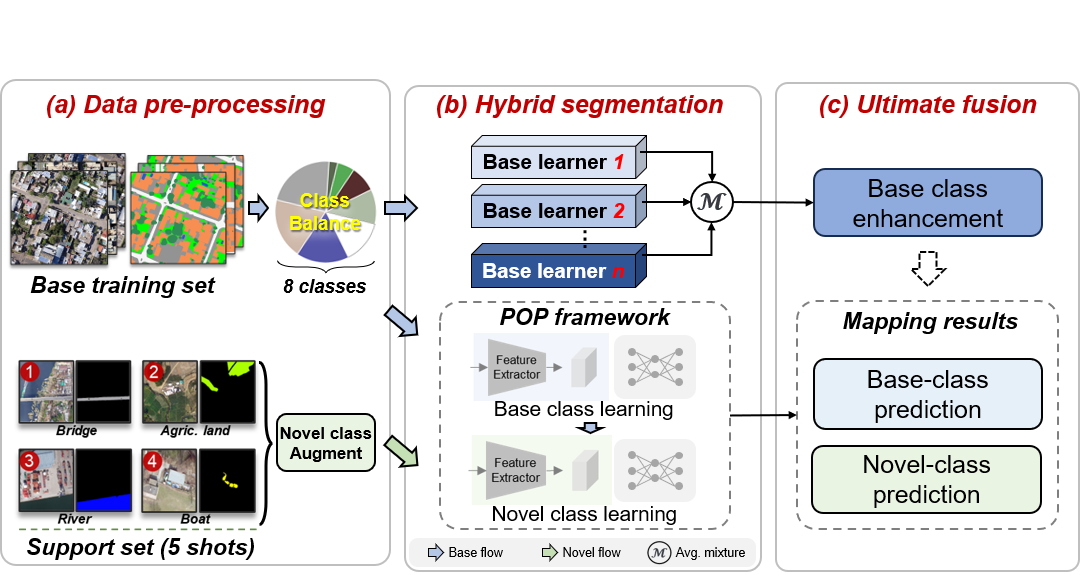}
    \caption{Overall flowchart of the GFSS-based land-cover mapping framework (SegLand), containing three main parts: (a) Data pre-processing, (b) Hybrid segmentation, and (c) Ultimate fusion.}
    \label{overall}
\end{figure*}

\subsection{Data analysis and pre-processing}


\begin{figure}[]
\centering
\includegraphics[width=0.8\linewidth]{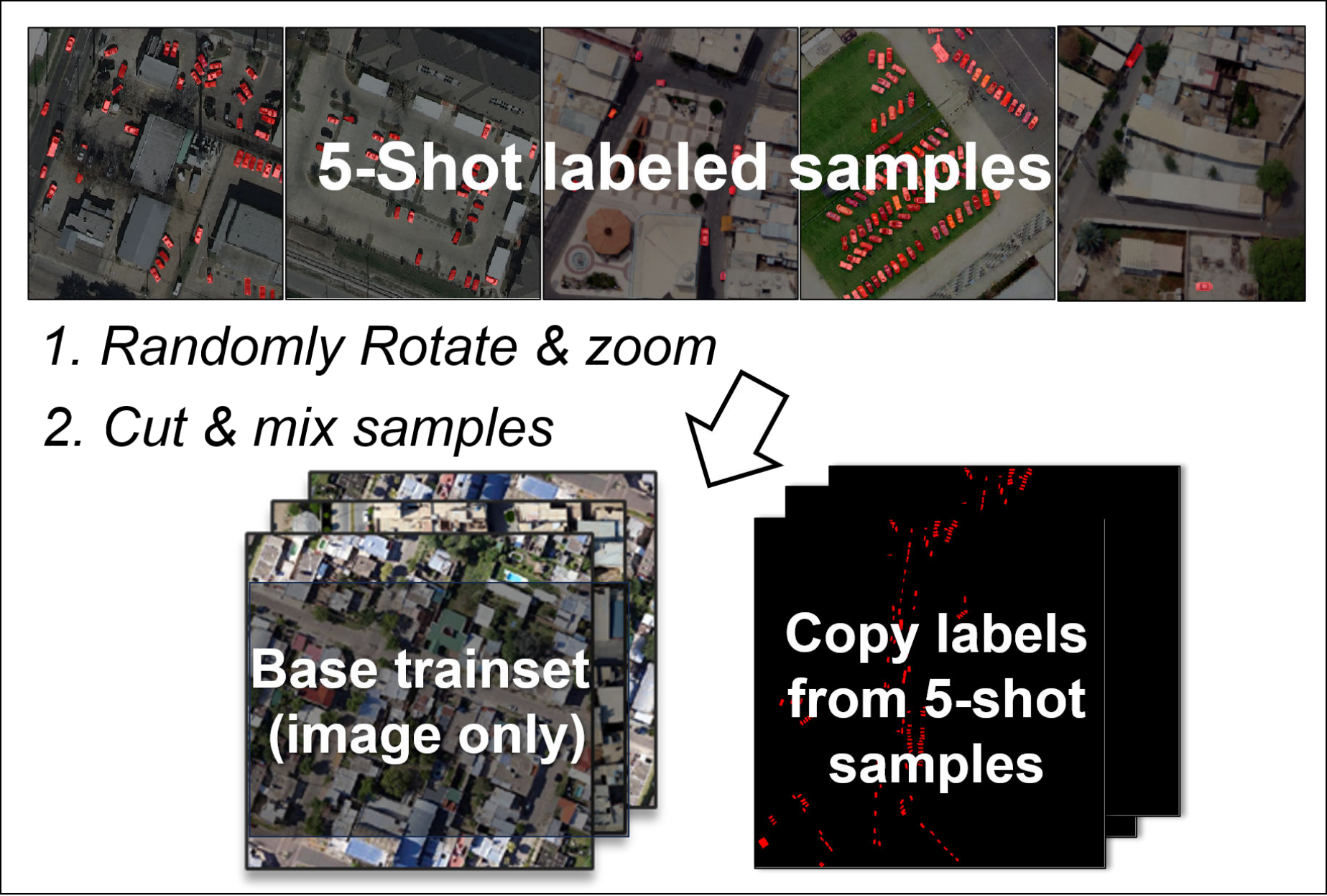}
\caption{Illustration of CutMix augmentation strategy. The images of the base training set are regarded as 'backgrounds' for novel classes.}
\label{cutmix_strategy}
\vspace{-1em}
\end{figure}	
\begin{figure}[]
\centering
\includegraphics[width=0.75\linewidth]{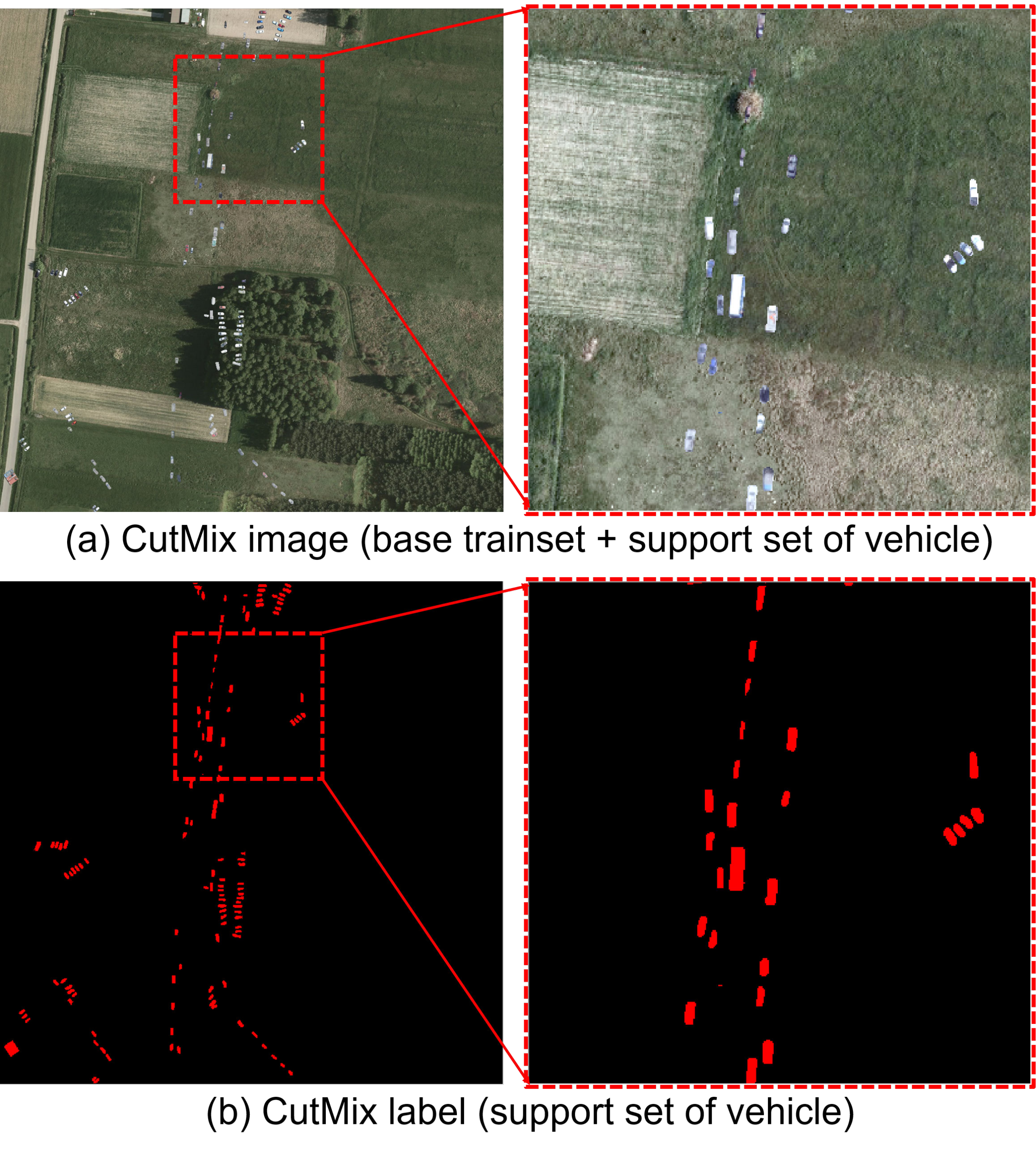}
\caption{Illustration of CutMix augmentation example. The novel class of vehicle is mixed in a random position of the image.}
\label{cutmix_sample}
\vspace{-1.5em}
\end{figure}	

The data preprocessing stage involves two primary operations. Firstly, we observed that the training set exhibits unbalanced distribution, with certain dominant classes containing the majority of examples, while a few classes, such as ``Sea, lake, pond'' and ``Bareland'', are represented by relatively few examples. Models trained on such data tend to perform poorly for weakly represented classes. To address this issue, we assign weights for different classes to devise a class-balanced loss.
We first calculate weights to be inversely proportional to the class frequency.
Additionally, we introduce a "smoothed" version that assigns weights to be inversely proportional to the square root of the class frequency \cite{cui2019class}. This simple heuristic method has been proven to be effective.

Secondly, we apply common data augmentation strategies such as random flips and crops to incorporate priors of invariance to translation and reflection to enhance performance. 
Furthermore, to enhance the proportion of samples representing novel classes within the training set, we have introduced a novel augmentation strategy called NovelCutMix based on CutMix \cite{yun2019cutmix}, as illustrated in Fig. \ref {cutmix_strategy}.
This strategy involves generating new training samples that exclusively represent novel classes by cutting and pasting patches between the validation set and the training set. 
Specifically, for each image from the validation set that only includes novel classes, we cut patches and paste them onto the corresponding regions of several images from the training set, and the ground truth labels of these training set images are replaced with the label of the validation set image.
Let $\mathbf{X} \in \mathbb{R}^{H \times W \times 3}$ and $\mathbf{Y}$ denote a image and its label, respectively. 
The goal of our proposed NovelCutMix is to generate a new training sample that only includes novel classes ($\tilde{\mathbf{X}}_T$, $\tilde{\mathbf{Y}}_T$) by replacing regions of a training sample ($\mathbf{X}_T$, $\mathbf{Y}_T$) with a validation sample ($\mathbf{X}_V$, $\mathbf{Y}_V$).
The replacement operation can be formulated as 
$$
\begin{aligned}
& \tilde{\mathbf{X}}_T=\mathbf{M} \odot \tilde{\mathbf{X}}_V+(\mathbf{1}-\mathbf{M}) \odot \mathbf{X}_T \\
& \tilde{\mathbf{Y}}_T=\mathbf{Y}_V,
\end{aligned}
$$
where $\mathbf{M} \in\{0,1\}^{H \times W}$ denotes a binary mask indicating where to drop out and fill in from two images, $\mathbf{1}$ is a binary mask filled with ones, and $\odot$ is element-wise multiplication.
The binary mask $\mathbf{M}$ indicating the regions where novel classes exist is determined by 

\begin{equation}
	\mathbf{M}^{i j}=\left\{\begin{array}{l}
			1, \text { if } \mathbf{Y}_V^{ij} \textgreater 0 \\
			0, \text { otherwise, }
		\end{array}\right. 
\end{equation}
where $i$ and $i$ represent the row and column of $\mathbf{M}$ or $\mathbf{Y}_V$.

As displayed in Fig. \ref {cutmix_sample}, the NovelCutMix can augment a sample of novel classes and generate a locally natural image, thereby enhancing the model's robustness against input corruptions and its ability to recognize novel classes.
\begin{figure*}[!h]
    \centering
    \includegraphics[width=0.95\linewidth]{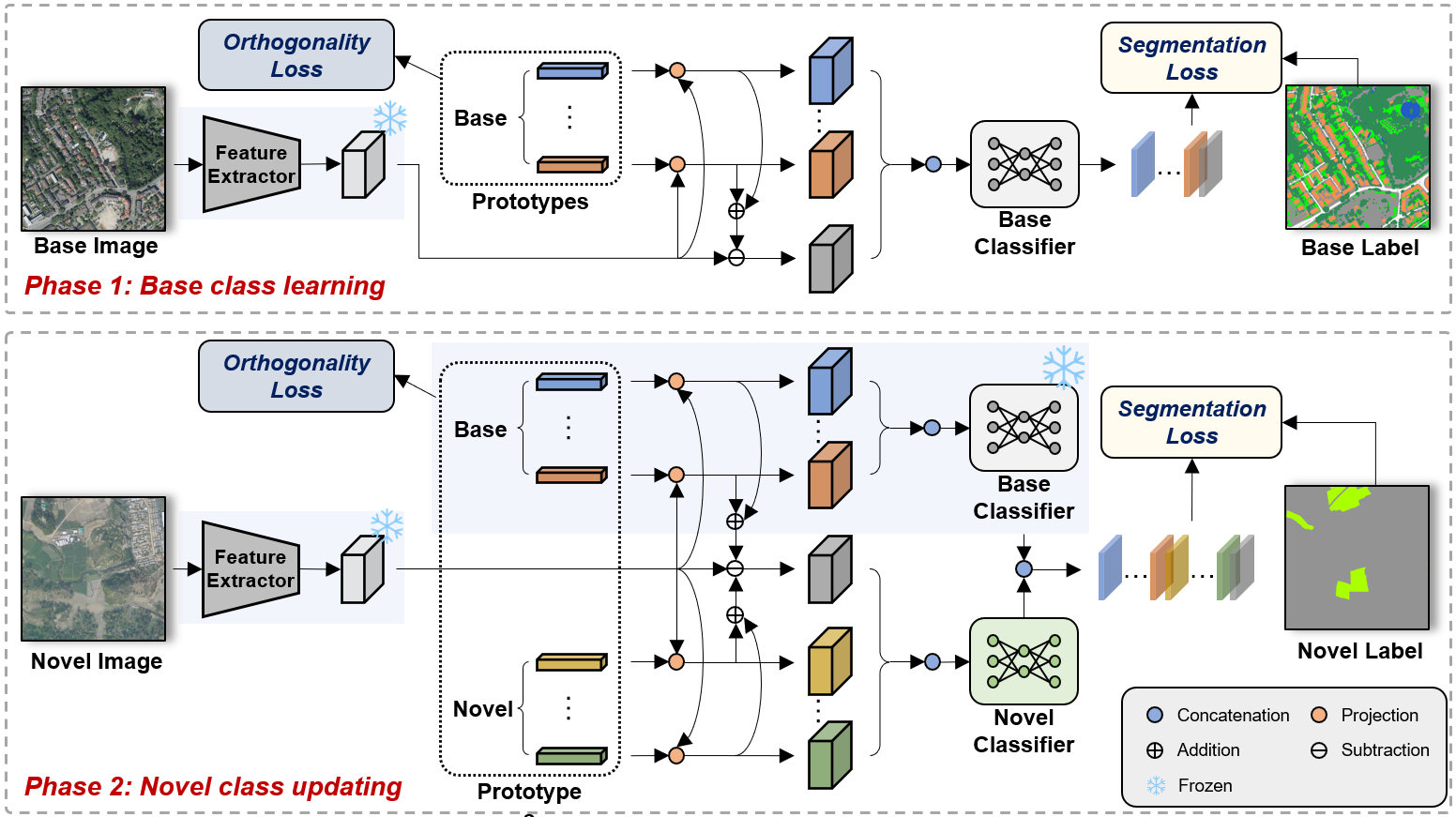}
    \caption{The overall framework of the adopted POP framework for GFSS, which consists of two phases: (a) Base class learning, (b) Novel class updating.}
    \vspace{-1em}
    \label{POP}
\end{figure*}
\subsection{Hybrid segmentation}
\subsubsection{Multiple base learner}
In order to improve model performance in recognizing base classes to the full extent, we first train multiple advanced baseline models separately using the base training set.
The HRNet48 \cite{9052469}, ResNeXt101 \cite{xie2017aggregated}, EfficientNetb7 \cite{tan2019efficientnet}, and UNetFormer \cite{wang2022unetformer} are selected as the baseline models based on their state-of-the-art performance in general segmentation task and land-cover classification task, each offering distinct network structures capable of learning diverse feature patterns.
Subsequently, an average fusion operation is employed to softly combine the predicted results from these four models.
This operation averages the probabilities calculated by each individual network, thereby ensuring the retention of high-confidence common parts in all prediction results and enhancing the ensemble's performance to outperform the best individual baseline models.
\vspace{-1em}
\subsubsection{Projection onto Orthogonal Prototypes Network}
\vspace{-0.5em}
In this part, a recent framework for generalized few-shot segmentation, called Projection onto Orthogonal Prototypes (POP) is introduced to explore novel classes in remote sensing images. As mentioned in Section 2.1, FSS mainly focuses on learning the prototypes for each class and classifying each query by measuring the similarity of queries and prototypes. It can be easily indicated that the accuracy of the classification heavily depends on the learning of prototypes. When queries of novel classes are added for prototype learning, the FSS methods fail to maintain the segmentation accuracy of the base classes while precisely recognizing the novel classes. To solve this problem, GFSS methods (introduced in Section 2.1) are proposed to balance the trade-off between extracting the base and novel classes. In order to learn the representation for both base and novel classes, the GFSS learning pipeline is generally divided into two stages. The first stage focuses on the base class learning. By training the parameters of the base class prototypes and the base classifier, the model extracts prototypes for the base classes and stores them for the next stage. In the second stage, samples containing queries for the novel classes are sent to the model for updating the prototype bank. Under the 2-stage training paradigm, the features of the base classes are sufficiently extracted, while the novel classes can be consistently updated whenever there are new samples to be learned.
\begin{figure}[t]
{
    \begin{minipage}[b]{\hsize}
     \centering
    \includegraphics[width=\linewidth]{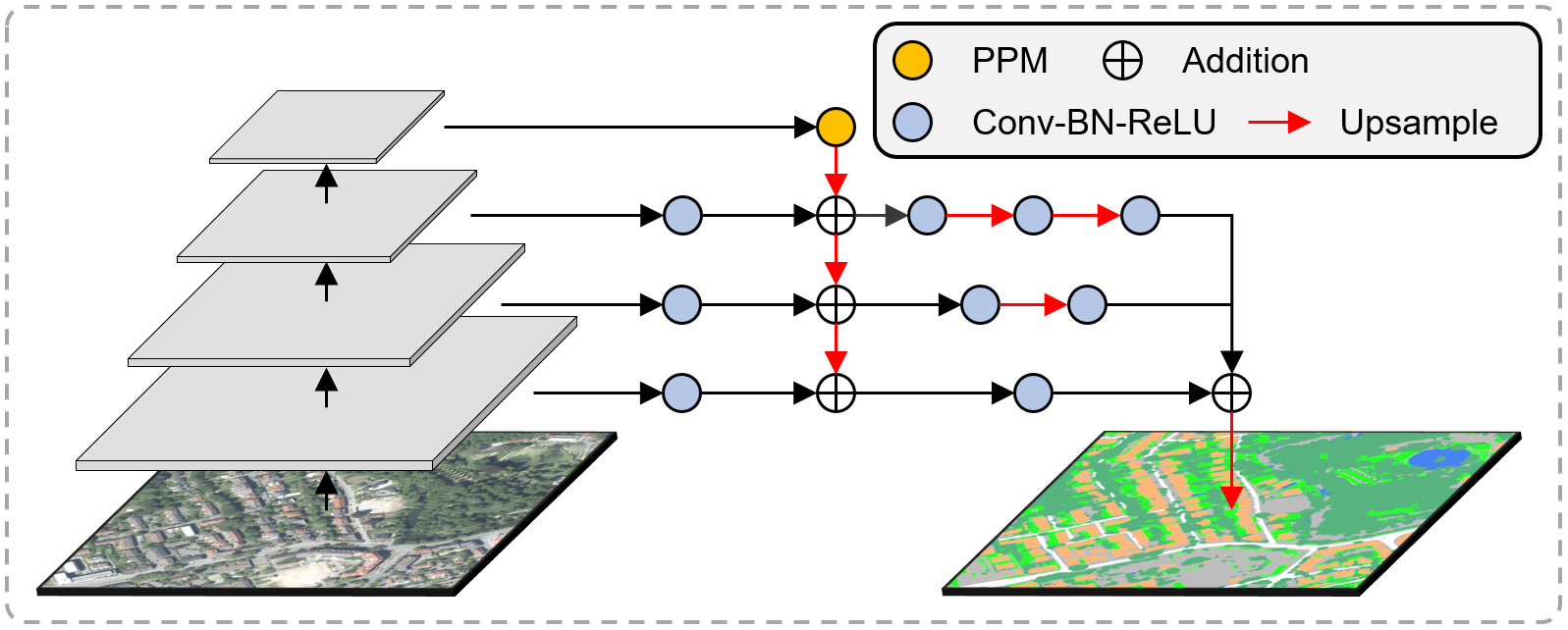}
    \end{minipage}
    } 
    \vspace{-2em}
\caption{The structure of the proposed UperNetPlus decoder.}
\label{decoder}
\vspace{-1em}
\end{figure}

Apparently, the two stages of the GFSS training paradigm should not interfere with each other to reach high classification accuracy for both base and novel classes. The past GFSS methods proposed various strategies to guarantee the aforementioned prerequisite. As the state-of-the-art method of GFSS, projection onto the orthogonal prototypes framework (POP) \cite{liu2023learning} ensures the non-interference of the 2-stage training by bringing the orthogonality into the training paradigm. As illustrated in Fig. \ref{POP}, the POP framework puts forward two solutions for the ensuring of the non-interference:
\begin{enumerate}
    \item \emph{The learned base prototypes and base classifier are frozen in the second stage to make sure that the backward of gradients will not affect the parameters of the base class learning part;}
    \item \emph{Only the background features are excluded from the frozen part for that the novel classes are extracted in the background;}
    \item \emph{Except for the basic segmentation loss for constraining the difference between segmentation results and the labels, the orthogonality loss is brought in to keep the orthogonality of the learned prototypes, i.e. generating a set of orthogonal basis of the feature space. The classification of each class (including both base and novel classes) under the orthogonal basis will not influence each other.}  
\end{enumerate}

With the above solutions, the POP framework shows superiority over the past GFSS frameworks and for this reason, it is adopted as our basic framework to solve the GFSS task of remote sensing. However, there are some modifications for training settings during the adoption of POP:
\begin{enumerate}
    \item \emph{Considering that the dataset released in the competition is much tiny compared with the pre-trained dataset (e.g. ImageNet\cite{deng2009imagenet}, Coco\cite{lin2014microsoft}), the feature extractor is also frozen compared with the original POP framework, in order to avoid overfitting;}
    \item \emph{The provided novel samples are only labeled with novel classes and the base classes in the samples are all annotated as background, which results in the confusion of the definition of background in both base and novel samples. To overcome the situation, the background of novel samples is ignored in the second stage of POP for steady novel class updating.} 
\end{enumerate}

Except for the above modifications, the feature extractor in the POP framework is also explored and compared among all the prevalent backbones and decoders for better segmentation performance. With substantial experiments, several backbones (including ResNet\cite{he2016deep}, HRNet\cite{wang2020deep}, SwinTransformer\cite{liu2021swin}, ConvNext\cite{liu2022convnet} and LSKNet\cite{li2023large}) and decoders (including FPN\cite{kirillov2019panoptic}, FarSeg\cite{zheng2020foreground}, UperNet\cite{xiao2018unified}) are compared for the optimal feature extractor composition. For the selection of backbone, Swin Transformer stands out among all the options, for its powerful ability of local-correlation modeling and detail maintaining. As for the decoder, a simple decoder called UperNetPlus (demonstrated in Fig. \ref{decoder}) is proposed here for feature decoding, which is simply composed of a simple panoptic-fpn structure and a pyramid pooling module (PPM\cite{zhao2017pyramid}). This decoder recovers the rich details of the segmentation results by applying a progressive upsampling strategy while avoiding large computations. With the powerful Swin Transformer as the backbone and UperNetPlus as the decoder, our proposed feature extractor achieves high accuracy in both base and novel class segmentation.
\begin{figure}[]
    \centering
    \includegraphics[width=\linewidth]{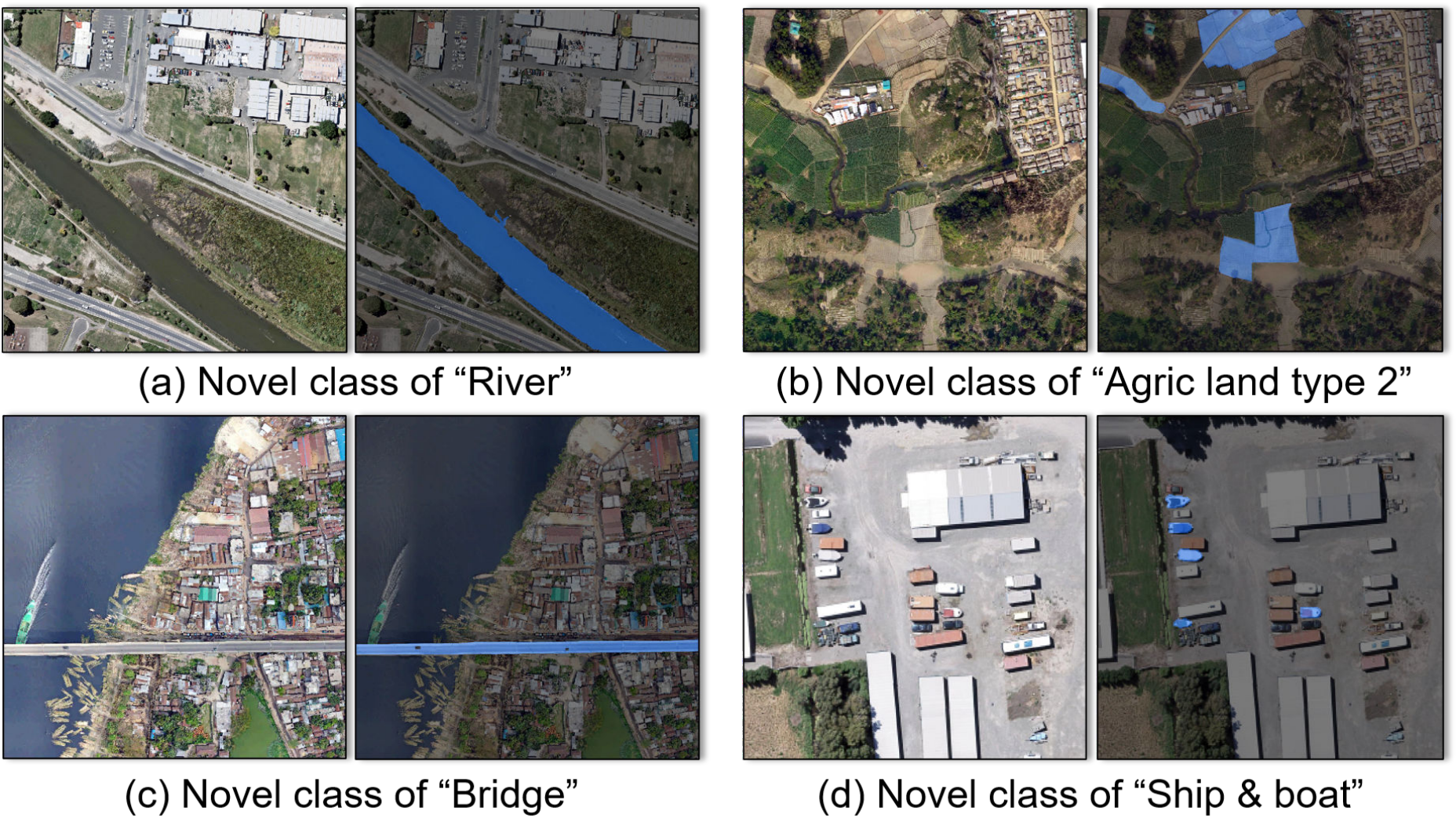}
    \caption{Sample of each novel class containing in the 5-shot support set (\textcolor[rgb]{0,0,0.8}{\textbf{Phase 1}}). The class of "Bridge" is named "Road type 2" in the dataset.}
    \label{novel-class1}
    \vspace{-1em}
\end{figure}
\begin{figure}[]
    \centering
    \includegraphics[width=\linewidth]{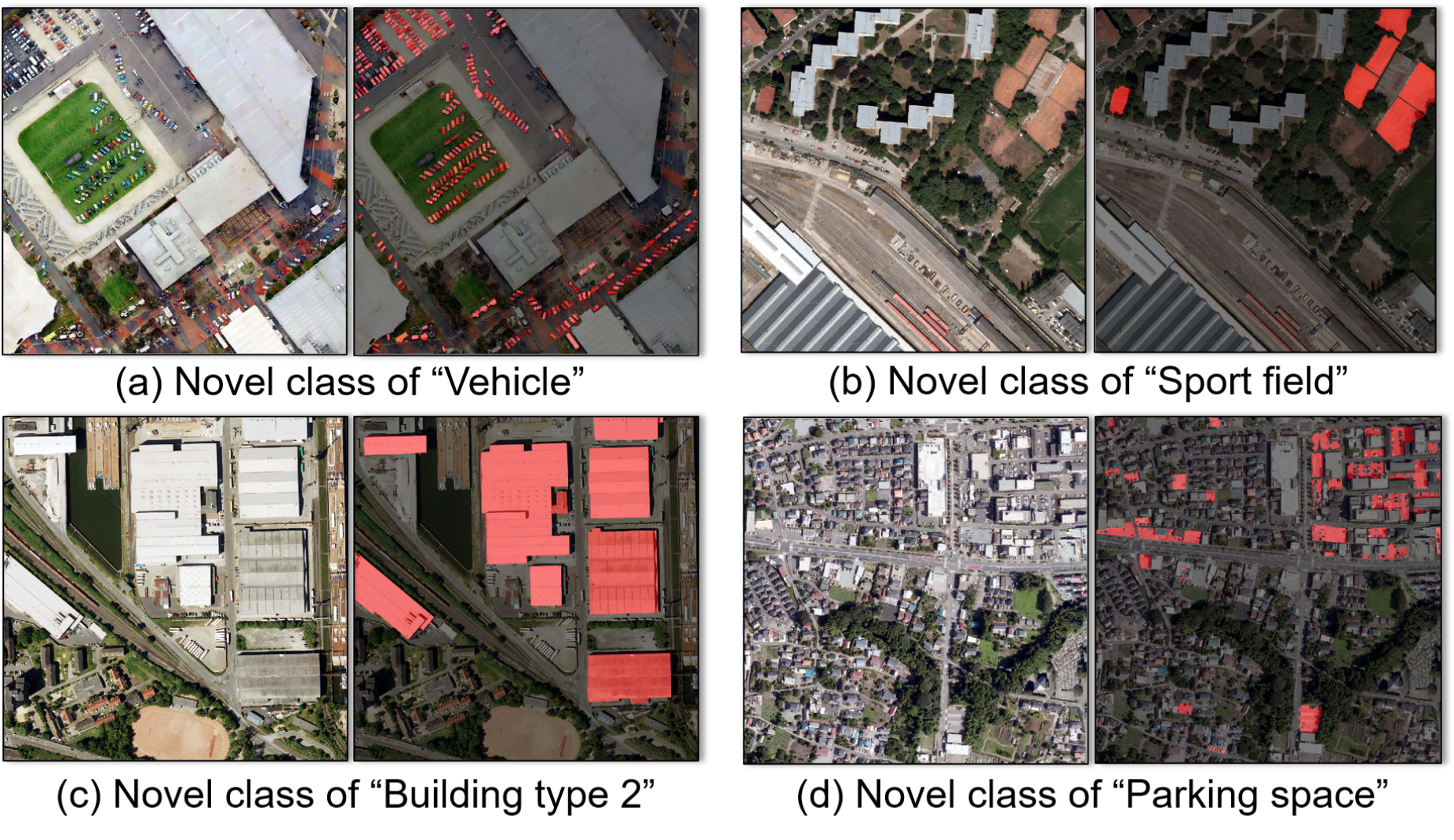}
    \caption{Sample of each novel class containing in the 5-shot support set (\textcolor[rgb]{0.8,0,0}{\textbf{Phase 2}}). }
    \label{novel-class2}
        \vspace{-1em}
\end{figure}
\subsection{Ultimate Fusion}
In this part, we make the ultimate fusion at the decision level by synthetically considering the overall results of base learners and the POP network. On the one hand, since multiple base learners only trained with the base training set, the land objects are segmented into the eight base classes shown in Fig. \ref{base visual}. Moreover, as we use an Avg. mixture to integrate the four model's results, the produced land-cover maps with base classes can be seen as a stable reference. On the other hand, the POP network simultaneously predicted the base and novel land objects that appeared on the images. To enhance the segmentation results of base classes, we use the mapping results of base learners to constrain the base-class prediction of the POP network. Specifically, we conduct intersection and morphological operations between the above-mentioned results to reasonably optimize the base results. Similar strategies have been widely applied in related competitions to improve the final results \cite{li2022multi,10431324}.

%% file: sec/4_Experiment.tex
\section{Experiment}
\subsection{Experiment setting and evaluation}
All the experiments and evaluations are conducted in the OpenEarthMap Land Cover Mapping Few-Shot Challenge\footnote{\url{https://codalab.lisn.upsaclay.fr/competitions/17568\#results}}, which is also a part of the 3rd Workshop on Learning with Limited Labeled Data for Image and Video Understanding (L3D-IVU) in conjunction with the CVPR 2024 Conference. All training data is provided by the challenge organizers. In phase 1 of the challenge, 258 tiles (with a size of 512 $\times$ 512) of the base training set and 20 tiles of the support set (four novel classes $\times$ 5 shots shown in Fig. \ref{novel-class1}) are only utilized to conduct the framework training. In phase 2 of the challenge, a 5-shot support set of four different classes shown in  Fig. \ref{novel-class2} are provided.
\subsection{Base Learner Training}
\begin{table}[]
\centering
\begin{tabular}{l|ll}
\hline
Training data                      & Base learner   & mIoU (base) \\ \hline
\multirow{5}{*}{Base training set} & HRNet48 \cite{9052469}        & 47.88       \\
                                   & ResNeXt101 \cite{xie2017aggregated}    & 49.20       \\
                                   & EfficientNetb7 \cite{tan2019efficientnet} & 48.93       \\
                                   & Unetformer \cite{wang2022unetformer}     & \underline{49.80}       \\ \cline{2-3} 
                                   & Avg. mixture   & \textbf{51.10}       \\ \hline
\end{tabular}
\caption{Evaluation of several base learners used in hybrid segmentation structure. The Avg. mixture represents the results of four model fusion. The results are all from the challenge website.}
\label{base learner}
\end{table}

\begin{figure}
    \centering
\includegraphics[width=1\linewidth]{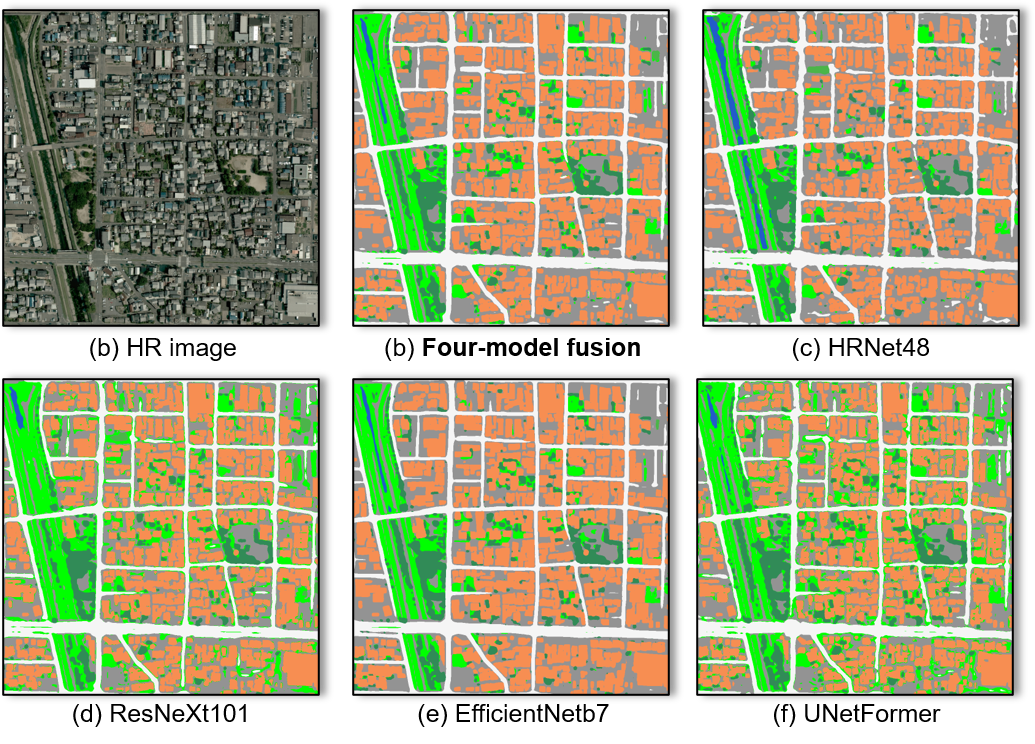}
    \vspace{-1.5em}
    \caption{Qualitative comparison of several base learners used in hybrid segmentation structure.}
    \label{base visual}
    \vspace{-0.5em}
\end{figure}

\begin{figure}
    \centering
\includegraphics[width=1\linewidth]{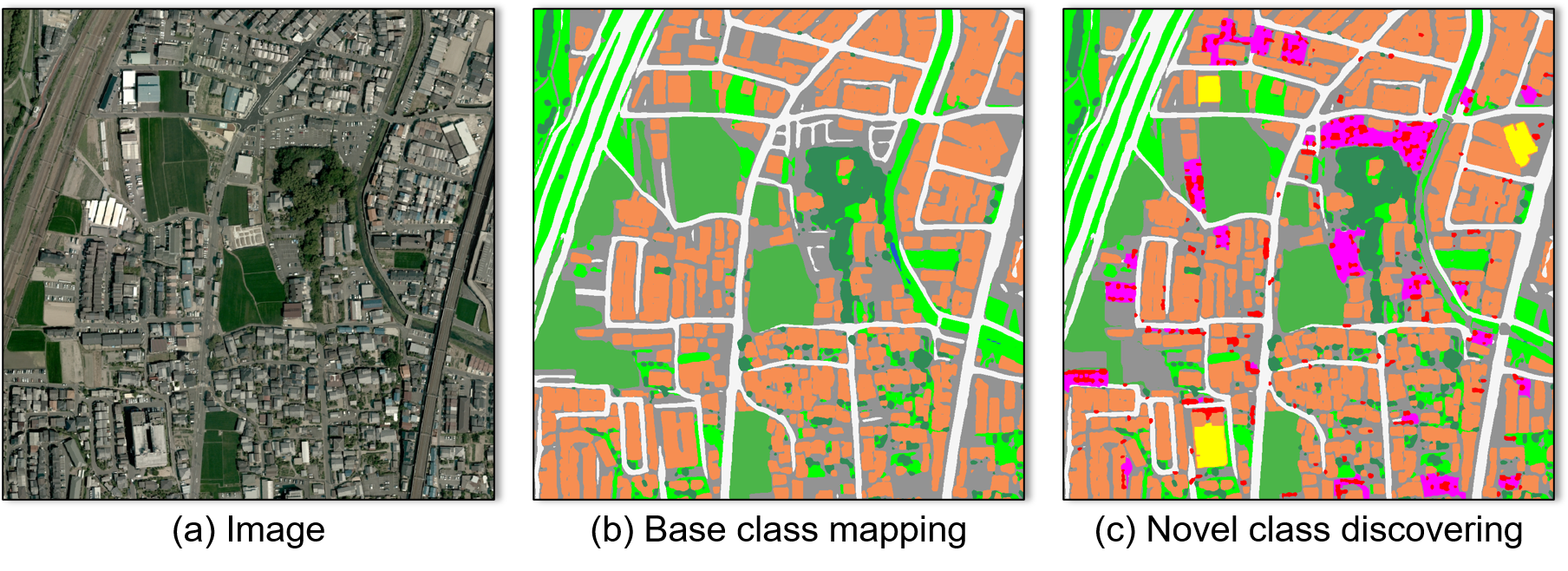}
    \vspace{-1.5em}
    \caption{An example of base-class land-cover mapping and further novel-class discovering.}
    \label{example}
    \vspace{-1em}
\end{figure}
Abundant experiments were conducted on the base training set for the selection of the optimal feature extractor in the GFSS training phase. As shown in Fig. \ref{overall} (b), four base models, including HRNet48, ResNeXt101, EfficientNetb7, and UNetFormer, are trained with only the base training set. Table. \ref{base learner} shows the mean intersection over union (mIoU) scores validated on the contest website. The UNetFormer, as a CNN-Transformer hybrid network \cite{wang2022unetformer},  shows the best performance on the base-class segmentation. By conducting the Avg. mixture operation, four base models are fused and obtain the highest mIoU (51.10\%) in the base classes. Fig. \ref{base visual} demonstrates the quantitative results of four base learners and the fusion land-cover map. The fusion map reveals a clearer traffic network and more accurate urban pattern corresponding to the HR image.
\subsection{GFSS Training}

\begin{figure*}[!h]
    \centering
\includegraphics[width=0.99\linewidth]{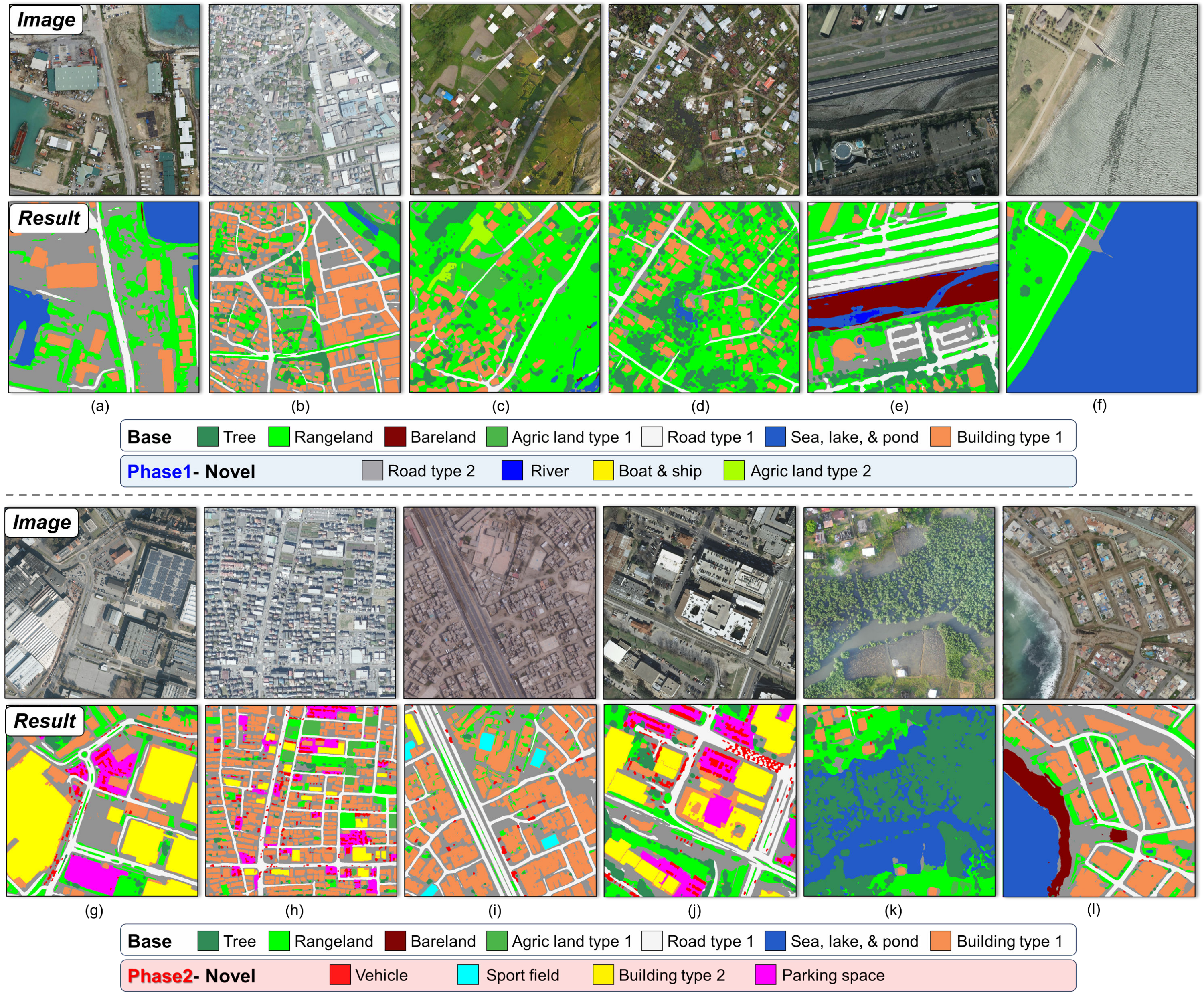}

    \caption{The visual results of final land-cover maps produced by the SegLand. The patches of (a–f) are sampled from the test images of challenge \textcolor[rgb]{0,0,0.8}{\textbf{phase 1}}. The patches of (g–l) are sampled from the test images of challenge \textcolor[rgb]{0.8,0,0}{\textbf{phase 2}}.}
    \label{novel-update}
    \vspace{-1em}
\end{figure*}
\begin{table}[]
\centering
\begin{tabular}{cllc}
\hline
\multicolumn{1}{l}{Backbone}    & Decoders    & mIoU  & \multicolumn{1}{l}{Epochs} \\ \hline
\multirow{4}{*}{Swin-T\cite{liu2021swin}} & FPN\cite{kirillov2019panoptic}         & 60.74 & \multirow{4}{*}{50}        \\
                                & FarSeg\cite{zheng2020foreground}      & 61.95 &                            \\
                                & UperNet\cite{xiao2018unified}     & \underline{62.21} &                            \\
                                & UperNetPlus & \textbf{64.56} &                            \\ \hline
\end{tabular}
\caption{Evaluation of SwinTransformer with different decoders used in the POP network. The results are all from a locally split test set (sampled 20\% data from the 258 base training set).}
\vspace{-0.5em}
\label{table2}
\end{table}

\begin{table}[]
\centering
\begin{tabular}{ll|ll}
\hline
Base class         & IoU   & Novel class          & IoU   \\ \hline
Tree               & 62.70 & \textcolor[rgb]{0,0,0.8}{\textbf{Road type 2 }}& 57.57  \\
Rangeland          & 55.52 & \textcolor[rgb]{0,0,0.8}{\textbf{River}}                & 10.87 \\
Bareland           & 42.79 & \textcolor[rgb]{0,0,0.8}{\textbf{Boat \& ship}}        & 57.06     \\
Agric land type 1  & 71.58 & \textcolor[rgb]{0,0,0.8}{\textbf{Agric land type 2}}    & 8.22  \\
Road type 1        & 59.29 &                      &       \\
Sea, lake, \& pond & 37.84 &                      &       \\
Building type 1    & 57.57 &                      &       \\ \hline
\textbf{mIoU (base) }       & \textbf{55.32} & \textbf{mIoU (novel)        } & \textbf{21.73} \\ \hline
\end{tabular}

\caption{The overall evaluation in \textcolor[rgb]{0,0,0.8}{\textbf{phase 1}} of the challenge produced by the proposed SegLand framework}
\label{overall quantitative results1}
\vspace{-0.5em}
\end{table}

\begin{table}[]
\centering
\begin{tabular}{ll|ll}
\hline
Base class         & IoU   & Novel class          & IoU   \\ \hline
Tree               & 69.18 & \textcolor[rgb]{0.8,0,0}{\textbf{Vehicle}} & 45.84  \\
Rangeland          & 53.03 & \textcolor[rgb]{0.8,0,0}{\textbf{Parking space }}               & 49.75 \\
Bareland           & 30.92 & \textcolor[rgb]{0.8,0,0}{\textbf{Sports field  }}       & 55.87     \\
Agric land type 1  & 62.30 & \textcolor[rgb]{0.8,0,0}{\textbf{Building type 2}}    & 61.92  \\
Road type 1        & 63.74 &                      &       \\
Sea, lake, \& pond & 53.27 &                      &       \\
Building type 1    & 61.74 &                      &       \\ \hline
\textbf{mIoU (base) }       & \textbf{56.27} & \textbf{mIoU (novel)        } & \textbf{53.34} \\ \hline
\end{tabular}

\caption{The overall evaluation in \textcolor[rgb]{0.8,0,0}{\textbf{phase 2}} of the challenge produced by the proposed SegLand framework}
\label{overall quantitative results2}
\vspace{-0.5em}
\end{table}

\begin{table}
    \centering
    \begin{tabular}{c|l|c}
    \hline
      Rank   & Team   & Total mIoU\\
      \hline
        1 & \textit{AsheLee} \textbf{(our)}& 54.52\\
         2& \textit{earth-insights} & 47.02\\
        3 & \textit{tiantian} & 44.06\\
         4&  \textit{yyc}& 42.73\\
         \hline
    \end{tabular}
    \caption{The final leaderboard (\textcolor[rgb]{0.8,0,0}{\textbf{phase 2}}) of the challenges. The total mIoU is $0.4\times \text{base mIoU} + 0.6\times \text{novel mIoU}$.}
    \label{leaderborad}
    \vspace{-1em}
\end{table}
Experiments conducted with a 5-shot dataset are illustrated to evaluate the effectiveness of the adopted POP framework for the GFSS. The first part of the experiments focuses on the validation of decoders. The results illustrated in Table \ref{table2} reveal the significant improvements (2.35\% mIoU) of UperNetPlus over UperNet, which shows that the FPN-style progressive upsampling strategy has better performance in remote sensing image feature extraction.

The second part is the novel class updating. After the base class learning phase, the learned feature extractor and base prototypes are frozen while novel samples are sent to the model for novel class learning. Fig. \ref{example} shows an example of base-class learning and novel-class updating results. From the results of phases 1 and 2 shown in Fig. \ref{novel-update}, Tables \ref{overall quantitative results1}, and \ref{overall quantitative results2}, the segmentation accuracy of the base classes is promising, and the novel classes are also successfully separated from the background and identified. The final results shown in the leaderboard (Table. \ref{leaderborad}) reveal the effectiveness of the proposed SegLand framework in solving GFSS tasks of remote-sensing data.

%% file: sec/5_Conclusion.tex
\vspace{-0.5em}
\section{Conclusion}
In this paper, a GFSS-based land-cover mapping framework is proposed to discover the novel land-cover classes that appear on the base land-cover maps. 
By considering the various scales of complex land objects and the interconnection between base and novel classes, the proposed SegLand framework is designed with three parts (a) Data pre-processing, (b) Hybrid segmentation structure, and (c) Ultimate fusion.
Experiments in the challenge show two findings: (1) The modified POP network is able to learn and update novel classes that appear in land cover mapping from a small number of labeled data. (2) The SegLand framework combines the hybrid segmentation mode and ultimate fusion process to produce land-cover mapping results by stably learning base class and accurately spot novel classes. In general, SegLand shows the potential to contextualize generalized few-shot segmentation and large-scale HR land-cover mapping and further facilitate many downstream Earth observation applications.
\vspace{+1em}
\\
\textbf{\large Acknowledgments}\vspace{+0.5em}
\\
This work has been supported by the National Key Research and Development Program of China (grant no. 2022YFB3903605) and the National Natural Science Foundation of China (grant no.42071322).